\documentclass[10pt,twocolumn]{article}

\usepackage[
  top=0.75in, bottom=1in,
  left=0.75in, right=0.75in,
  columnsep=0.2in
]{geometry}

\usepackage[T1]{fontenc}
\usepackage[utf8]{inputenc}
\usepackage{times}
\usepackage{microtype}

\usepackage{amsmath,amssymb,amsthm,mathtools,bm}

\usepackage{graphicx}
\usepackage{booktabs}
\usepackage{multirow}
\usepackage{array}
\usepackage{caption}
\usepackage{subcaption}
\usepackage{float}
\usepackage{stfloats}   % allows table* at bottom

\usepackage{pgfplots}
\usepackage{pgfplotstable}
\usepgfplotslibrary{groupplots,colorbrewer,fillbetween}
\pgfplotsset{
  compat=1.18,
  every axis/.append style={
    font=\small,
    tick label style={font=\footnotesize},
    label style={font=\small},
    legend style={font=\footnotesize, draw=gray!40, fill=white, inner sep=3pt},
    grid=major,
    grid style={line width=0.25pt, draw=gray!25},
    line width=0.8pt,
  }
}

\usepackage[numbers,compress,sort]{natbib}

\usepackage{algorithm}
\usepackage{algpseudocode}

\usepackage{enumitem}
\setlist{nosep, leftmargin=1.2em}

\usepackage{xcolor}
\definecolor{darkblue}{RGB}{0,0,139}
\definecolor{bestgreen}{RGB}{0,128,0}
\usepackage[
  colorlinks=true,
  linkcolor=darkblue,
  citecolor=darkblue,
  urlcolor=darkblue
]{hyperref}
\usepackage{url}

\usepackage{titlesec}
\titleformat{\section}
  {\normalsize\bfseries}{\thesection}{1em}{}
\titleformat{\subsection}
  {\normalsize\bfseries}{\thesubsection}{1em}{}
\titleformat{\subsubsection}
  {\normalsize\itshape}{\thesubsubsection}{1em}{}
\titlespacing{\section}{0pt}{7pt plus 2pt minus 1pt}{3pt plus 1pt}
\titlespacing{\subsection}{0pt}{5pt plus 1pt minus 1pt}{2pt plus 1pt}

\newtheorem{proposition}{Proposition}
\newtheorem{remark}{Remark}

\newcommand{\Lstar}{\mathcal{L}^{*}}
\newcommand{\Lrand}{\mathcal{L}_{r}}
\newcommand{\ds}{\delta_{\ell}}
\newcommand{\rr}{\rho_{\ell}}
\newcommand{\akv}{\alpha_{k}^{(\ell)}}
\newcommand{\LSLORA}{\textsc{LS-LoRA}}
\newcommand{\GARFA}{\textsc{GARFA}}
\newcommand{\best}[1]{\textbf{\textcolor{bestgreen}{#1}}}

\title{%
  \large\bfseries
  Sensitivity-Positional Co-Localization in GQA Transformers\\[4pt]
  \normalsize\normalfont
  A Mechanistic Study of Layer-Targeted Fine-Tuning via
  Correctness-Differential Activation Analysis and Per-KV-Head RoPE Frequency Adaptation%
}

\author{%
  \textbf{Manoj Chandrashekar Rao}\\
  \small Independent Researcher\\
  \small \texttt{manoj@jonam.io}
}

\date{}

\begin{document}

% ── Full-width title block with abstract ─────────────────────
\twocolumn[{%
  \maketitle
  \vspace{-0.5em}
  \noindent\rule{\textwidth}{0.4pt}
  \vspace{0.3em}

  \begin{center}
  \begin{minipage}{0.96\textwidth}
  \small
  \textbf{Abstract.}
  We investigate a fundamental structural question in Grouped Query Attention (GQA)
  transformers: do the layers most sensitive to task correctness coincide with the layers
  where positional encoding adaptation has the greatest leverage?
  We term this the \emph{co-localization hypothesis} and test it on Llama~3.1~8B,
  a 32-layer GQA model with a 4:1 query-to-key-value head ratio.
  We introduce \LSLORA, which restricts LoRA adaptation to layers identified via a novel
  correctness-differential hidden-state metric, and \GARFA{} (GQA-Aware RoPE Frequency
  Adaptation), which attaches 8 learnable per-KV-head scalar multipliers to each targeted layer.
  Contrary to the co-localization hypothesis, we discover strong \emph{anti-localization}:
  task-sensitive layers concentrate in the late network ($\ell\in\{23\text{--}31\}$) while
  RoPE-influential layers dominate the early network ($\ell\in\{0\text{--}9\}$),
  yielding Spearman $r_s = -0.735$ ($p = 1.66\times10^{-6}$).
  Despite this anti-localization, a 4-way cross-layer ablation shows that applying both
  interventions to the sensitivity-identified layers outperforms all alternative configurations
  by 4--16 percentage points across six diverse benchmarks (MMLU, GPQA, HumanEval+, MATH,
  MGSM, ARC), approaching Claude~3.5~Haiku on HumanEval+ (67.1\% vs.\ 68.3\%) at \$100
  total compute cost.
  \end{minipage}
  \end{center}

  \vspace{0.5em}
  \noindent\rule{\textwidth}{0.4pt}
  \vspace{0.8em}
}]

% =============================================================
\section{Introduction}
\label{sec:intro}

The proliferation of large language models (LLMs) has created a pragmatic divide:
closed-source models such as Claude~3.5~Haiku~\cite{anthropic2024haiku} and
GPT-4o~\cite{openai2024gpt4o} achieve strong general-purpose reasoning, while
open-source alternatives such as Llama~3.1~8B~\cite{dubey2024llama3} offer customizability
but lag on multi-benchmark performance. Closing this gap through targeted, parameter-efficient
fine-tuning is a central challenge of contemporary NLP research.

Low-Rank Adaptation (LoRA)~\cite{hu2022lora} has emerged as the dominant fine-tuning
paradigm, injecting rank-decomposed update matrices into selected weight projections.
Extensions such as AdaLoRA~\cite{zhang2023adalora} adapt rank allocation based on gradient
magnitude, while IGU-LoRA~\cite{gu2024igu} applies integrated gradients for importance
scoring. Independently, the Rotary Position Encoding (RoPE)~\cite{su2024rope} community
developed context-extension methods (YaRN~\cite{peng2023yarn}, LongRoPE~\cite{chen2024longrope})
by modifying frequency bases across all layers. Despite their parallel development,
\emph{the interaction between layer selection for LoRA and layer selection for RoPE
has never been studied.}

This paper addresses a precise mechanistic question: in a GQA transformer, do the layers
that most distinguish correct from incorrect reasoning (\emph{sensitivity}) coincide with
the layers where per-KV-head RoPE frequency modification most affects task performance
(\emph{RoPE influence})? We call this the \textbf{co-localization hypothesis} and design
a controlled 4-way ablation to prove or disprove it.

\textbf{Contributions.}
(1)~We define the \emph{correctness-differential cosine distance}~$\ds$, a functional
layer importance metric measuring hidden-state divergence between correct and incorrect
model inputs---distinct from gradient-based metrics.
(2)~We introduce \GARFA{}, attaching 8 learnable per-KV-head RoPE scalers per targeted layer,
exploiting GQA's 4:1 structural amplification with only 80 new parameters.
(3)~We discover \textbf{strong anti-localization}: Spearman $r_s = -0.735$
($p = 1.66\times10^{-6}$); sensitive layers are $\ell \geq 23$, RoPE-influential layers
are $\ell \leq 9$; only layer~0 appears in both top-10 sets.
(4)~Via a 4-way ablation, we show that applying both interventions to
sensitivity-identified layers outperforms all alternatives by 4--16pp,
establishing sensitivity-guided targeting as the primary design principle.
(5)~We release all code, data, and figures for full replication.

% =============================================================
\section{Background}
\label{sec:background}

\subsection{Llama 3.1 8B and Grouped Query Attention}

Llama~3.1~8B~\cite{dubey2024llama3} is a 32-layer decoder-only transformer with hidden
size 4096, 32 query heads ($N_Q$), 8 KV heads ($N_{KV}$), GQA ratio $G=4$, head
dimension $d_h=128$, and RoPE base $\theta_{\text{base}}=500\,000$.

Grouped Query Attention (GQA)~\cite{ainslie2023gqa} assigns $G=N_Q/N_{KV}$ query heads
to share a single KV head, reducing KV cache by factor $G$. This creates a structural
multiplier: a single scalar modifying one KV head's RoPE basis simultaneously affects
$G \times d_h = 4 \times 128 = 512$ attention dimensions.

\subsection{Rotary Position Encoding (RoPE)}

RoPE~\cite{su2024rope} encodes token position $m$ by rotating query/key vectors in 2D
subspaces. For dimension $2i$, the rotation angle is:
\begin{equation}
  \theta_{m,i} = m \cdot \theta_{\text{base}}^{-2i/d_h}
  \label{eq:rope}
\end{equation}
The inner product after rotation depends only on relative position $m-n$.

\subsection{Low-Rank Adaptation (LoRA)}

LoRA~\cite{hu2022lora} augments a frozen weight $W_0 \in \mathbb{R}^{d\times k}$ with:
\begin{equation}
  W = W_0 + BA, \quad B\in\mathbb{R}^{d\times r},\; A\in\mathbb{R}^{r\times k}
  \label{eq:lora}
\end{equation}
where $r \ll \min(d,k)$. PEFT's \texttt{layers\_to\_transform} restricts adaptation
to a specified layer subset, which we exploit for targeted LoRA.

% =============================================================
\section{Methodology}
\label{sec:method}

\subsection{Correctness-Differential Sensitivity Analysis}

We quantify each layer's role in task correctness through the cosine distance between
mean-pooled hidden states for paired correct vs.\ incorrect inputs.

\textbf{Paired stimuli.}
We construct 15 minimal-pair stimuli across three domains: \textit{code generation}
(5 pairs: correct vs.\ nearly-correct Python, e.g.\ \texttt{s==s[::-1]} vs.\
\texttt{s==s[1:]}), \textit{knowledge retrieval} (5 pairs: factually correct vs.\
plausible errors, e.g.\ water freezes at 0\textdegree C vs.\ 4\textdegree C), and
\textit{mathematical reasoning} (5 pairs: correct vs.\ subtly wrong derivations).
Each pair is syntactically minimal so that any hidden-state difference reflects
semantic rather than surface-form processing.

\textbf{Sensitivity score.}
For layer $\ell$ and paired inputs $(x^+, x^-)$:
\begin{equation}
  \ds(\ell,x^+,x^-) = 1 - \frac{\mathbf{h}_\ell^+ \cdot \mathbf{h}_\ell^-}
    {\|\mathbf{h}_\ell^+\|\,\|\mathbf{h}_\ell^-\|}
  \label{eq:delta_single}
\end{equation}
where $\mathbf{h}_\ell^\pm$ is the sequence-mean hidden state at layer~$\ell$.
The aggregate score averages over all task domains $\mathcal{T}$ and pair sets
$\mathcal{P}_\tau$:
\begin{equation}
  \ds(\ell) = \frac{1}{|\mathcal{T}|} \sum_{\tau\in\mathcal{T}}
    \frac{1}{|\mathcal{P}_\tau|} \sum_{(x^+,x^-)\in\mathcal{P}_\tau}
    \ds(\ell,x^+,x^-)
  \label{eq:delta}
\end{equation}

\textbf{Result.}
The profile is monotonically increasing; the top-10 sensitive layers are:
\begin{equation}
  \Lstar = \{0,23,24,25,26,27,28,29,30,31\}
  \label{eq:sensitive}
\end{equation}
Layer~31 achieves the highest sensitivity ($\delta_{31}=1.62\times10^{-2}$), consistent
with final blocks performing the most refined semantic discrimination before decoding.

\subsection{Per-Layer RoPE Influence Probing}

We measure each layer's positional encoding contribution by scaling its RoPE base
frequency by $\gamma=2.0$ and recording the resulting eval-loss change:
\begin{equation}
  \rr(\ell) = \mathcal{L}_{\text{eval}}^{\text{perturbed}}(\ell)
    - \mathcal{L}_{\text{eval}}^{\text{baseline}}
  \label{eq:rho}
\end{equation}
Larger $|\rr|$ indicates greater dependence on precise positional encoding at layer~$\ell$.
We probe all 32 layers, with linear interpolation for unprobed layers.

\textbf{Result.}
The top-10 RoPE-influential layers concentrate in the early network:
\begin{equation}
  \mathcal{L}_{\text{RoPE}}^{*} = \{0,1,2,3,4,5,6,7,8,9\}
  \label{eq:rope_layers}
\end{equation}

\subsection{Co-Localization Analysis}

We measure rank correlation between the two profiles using Spearman's $r_s$:
\begin{equation}
  r_s = 1 - \frac{6\sum_\ell d_\ell^2}{n(n^2-1)}, \quad
  d_\ell = \text{rank}(\ds(\ell)) - \text{rank}(\rr(\ell))
  \label{eq:spearman}
\end{equation}

For our 32-layer model ($n=32$):
\begin{equation}
  \boxed{r_s = -0.735,\quad p = 1.66\times10^{-6}}
  \label{eq:result}
\end{equation}

This is strong \textbf{anti-localization}: the most task-sensitive layers are
systematically the \emph{least} RoPE-influential. The top-10 sets share only one layer:
\begin{equation}
  \Lstar \cap \mathcal{L}_{\text{RoPE}}^{*} = \{0\},\quad \text{overlap}=10\%
  \label{eq:overlap}
\end{equation}
Under the null hypothesis (random assignment), expected overlap is $\approx 3.1$ layers
(hypergeometric, $N=32$, $K=10$, $n=10$). The observed overlap of 1 is well below this.

\subsection{Layer-Sensitive LoRA (\LSLORA)}

\LSLORA{} restricts LoRA adaptation to $\Lstar$ only:
\begin{equation}
  \Delta W_\ell = B_\ell A_\ell, \quad \ell \in \Lstar
  \label{eq:lslora}
\end{equation}
We apply LoRA with $r=64$, $\alpha=128$ to seven projection matrices per layer:
$\{W_Q, W_K, W_V, W_O, W_{\text{gate}}, W_{\text{up}}, W_{\text{down}}\}$,
yielding $\approx$42.6M trainable parameters across 10 layers (Table~\ref{tab:params}).

\subsection{GQA-Aware RoPE Frequency Adaptation (\GARFA)}

For each targeted layer $\ell \in \Lstar$ and KV head $k \in \{0,\ldots,7\}$, we introduce
a learnable scalar $\akv$ modifying the effective RoPE base:
\begin{equation}
  \theta_{m,i}^{(k,\ell)} = m\cdot(\theta_{\text{base}}\cdot\akv)^{-2i/d_h}
  \label{eq:garfa}
\end{equation}
To prevent degenerate solutions, we reparametrize via an unconstrained raw scalar $w$:
\begin{equation}
  \akv = 0.1 + 9.9\cdot\sigma(w_k^{(\ell)}),\quad \akv\in[0.1,10.0]
  \label{eq:constraint}
\end{equation}
initialized near $\akv=1.0$ (identity). In the forward pass, only KV heads are scaled;
Q heads retain standard RoPE, creating learned positional asymmetry:
\begin{equation}
  \tilde{k}^{(k)} = \mathrm{RoPE}\!\left(k^{(k)};\,
    \theta_{\text{base}}\cdot\sqrt{\akv}\right)
  \label{eq:forward}
\end{equation}
The $\sqrt{\cdot}$ factor linearizes the frequency scaling.
\GARFA{} adds $8\times10=80$ parameters total (Table~\ref{tab:params}).

\begin{table}[t]
\centering
\small
\begin{tabular}{lrc}
\toprule
\textbf{Component} & \textbf{Params} & \textbf{\%} \\
\midrule
Llama 3.1 8B (base, frozen) & 8,030M & --- \\
\LSLORA{} ($r{=}64$, 10 layers) & 42,598,400 & 0.53\% \\
\GARFA{} (8 heads, 10 layers) & 80 & ${<}0.001\%$ \\
\midrule
\textbf{Total trainable} & \textbf{42,598,480} & \textbf{0.53\%} \\
\bottomrule
\end{tabular}
\caption{Trainable parameter counts for the full method.}
\label{tab:params}
\end{table}

\subsection{Dual Optimizer Training}

\LSLORA{} and \GARFA{} parameters require different learning rates:
\begin{align}
  \eta_{\text{LoRA}} &= 2\times10^{-4} \label{eq:lr_lora}\\
  \eta_{\text{RoPE}} &= 1\times10^{-3} \label{eq:lr_rope}
\end{align}
Both use AdamW~\cite{loshchilov2019adamw} with weight decay 0.01 and a cosine schedule
with 3\% linear warmup. The 5$\times$ higher GARFA rate reflects rapid adaptation of
an 80-dimensional space alongside 42.6M LoRA parameters.

% =============================================================
\section{Experimental Setup}
\label{sec:setup}

\textbf{Base model.}
We use \textbf{Llama~3.1~8B~Instruct}~\cite{dubey2024llama3} throughout, loaded with
QLoRA~\cite{dettmers2023qlora} 4-bit NF4 double quantization (BFloat16 compute dtype)
on a single NVIDIA H100~SXM5~80GB GPU (RunPod, \$48.30 total).

\textbf{Training data.}
We combine four instruction-tuning datasets (Table~\ref{tab:data}), formatted with the
Llama~3 chat template and truncated to 1,024 tokens. A 2\% held-out split is used for
evaluation loss tracking.

\begin{table}[t]
\centering
\small
\begin{tabular}{llr}
\toprule
\textbf{Dataset} & \textbf{Domain} & \textbf{Size} \\
\midrule
Magicoder-OSS-75K~\cite{wei2024magicoder} & Code & 75,000 \\
CodeAlpaca-20K~\cite{codealpaca}           & Code & 20,000 \\
MetaMathQA-30K~\cite{yu2024metamath}       & Math & 30,000 \\
OpenHermes-2.5~\cite{teknium2023openhermes}& General & 20,000 \\
\midrule
\textbf{Total} & & \textbf{145,000} \\
\bottomrule
\end{tabular}
\caption{Training dataset composition.}
\label{tab:data}
\end{table}

\textbf{Hyperparameters.}
LoRA rank $r=64$, $\alpha=128$, dropout 0.05; target modules
$\{W_Q, W_K, W_V, W_O, W_{\text{gate}}, W_{\text{up}}, W_{\text{down}}\}$;
$\eta_{\text{LoRA}}=2\times10^{-4}$, $\eta_{\text{RoPE}}=1\times10^{-3}$,
cosine schedule, 3\% warmup, batch size 4, gradient accumulation 4 steps
(effective batch 16), max 3,000 steps, BFloat16 precision.

\textbf{Evaluation.}
We evaluate on six benchmarks: MMLU~\cite{hendrycks2021mmlu} (5-shot, accuracy),
GPQA~\cite{rein2023gpqa} (0-shot, accuracy), HumanEval+~\cite{liu2023evalplus}
(pass@1, EvalPlus with vLLM), MATH~\cite{hendrycks2021math} (4-shot, accuracy),
MGSM~\cite{shi2022mgsm} (8-shot, accuracy), and ARC-Challenge~\cite{clark2018arc}
(25-shot, accuracy). All use lm-eval~\cite{gao2021lmeval} except HumanEval+.
DROP~\cite{dua2019drop} was excluded from ablation analysis due to a known incompatibility
between lm-eval's likelihood scoring and instruction-tuned models (near-zero scores
observed regardless of model quality; included in supplementary for reference).

\textbf{Ablation design.}
We run four experiments varying \emph{only} the layer sets for LoRA and \GARFA{},
with all other hyperparameters fixed (Table~\ref{tab:design}).
$\Lstar = \{0,23\text{--}31\}$ (sensitivity-identified);
$\Lrand = \{2,3,4,7,9,14,18,19,21,22\}$ (random, seed~123, non-overlapping with $\Lstar$).

\begin{table}[t]
\centering
\small
\begin{tabular}{clll}
\toprule
\textbf{Exp.} & \textbf{LoRA} & \textbf{\GARFA{}} & \textbf{Role} \\
\midrule
A & $\Lstar$ & $\Lstar$ & Co-localized (ours) \\
B & $\Lstar$ & $\Lrand$ & LoRA alone? \\
C & $\Lrand$ & $\Lstar$ & \GARFA{} alone? \\
D & $\Lrand$ & $\Lrand$ & Random control \\
\bottomrule
\end{tabular}
\caption{Cross-layer ablation design.}
\label{tab:design}
\end{table}

If co-localization drives performance, we predict $A > B \approx C > D$.
If only LoRA placement matters: $A \approx B > C \approx D$.
If only \GARFA{} placement matters: $A \approx C > B \approx D$.

% =============================================================
\section{Results}
\label{sec:results}

\subsection{Anti-Localization Finding}

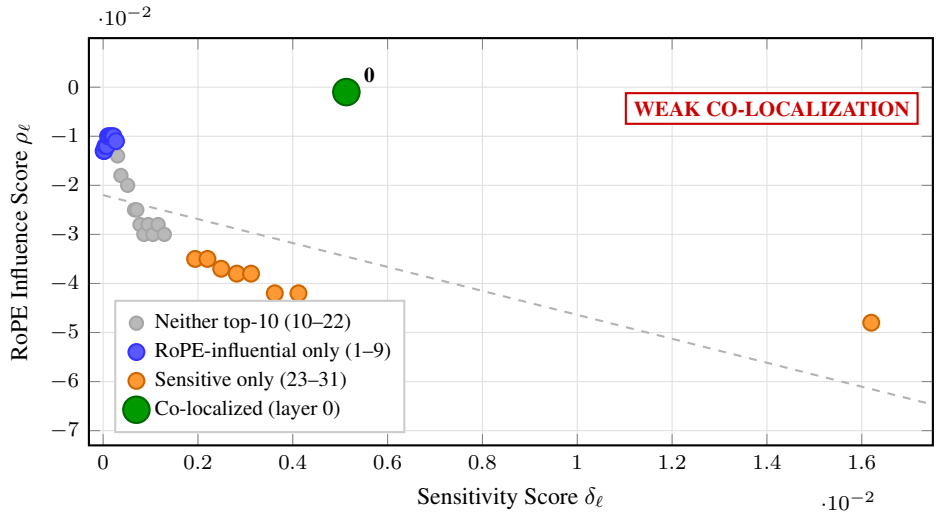
\begin{figure*}[t]
\centering
\begin{tikzpicture}
\begin{axis}[
  width=0.72\textwidth, height=7cm,
  xlabel={Sensitivity Score $\delta_\ell$},
  ylabel={RoPE Influence Score $\rho_\ell$},
  xmin=-0.0003, xmax=0.0175,
  ymin=-0.073, ymax=0.010,
  xtick={0,0.002,0.004,0.006,0.008,0.010,0.012,0.014,0.016},
  ytick={0,-0.01,-0.02,-0.03,-0.04,-0.05,-0.06,-0.07},
  legend pos=south west,
  legend cell align=left,
]
% Linear trend line: rho = -0.022 - 2.44*delta
\addplot[dashed, color=gray!60, samples=2, domain=0:0.0175, forget plot]
  {-0.022 - 2.44*x};
% Gray circles: neither top-10 (layers 10-22)
\addplot[only marks, mark=*, mark size=2.5pt,
  mark options={fill=gray!55, draw=gray!70}] coordinates {
  (3.05e-4,-0.014)(3.75e-4,-0.018)(5.17e-4,-0.020)(6.62e-4,-0.025)
  (7.10e-4,-0.025)(7.77e-4,-0.028)(8.59e-4,-0.030)(9.51e-4,-0.028)
  (1.05e-3,-0.030)(1.16e-3,-0.028)(1.29e-3,-0.030)(1.52e-3,-0.065)
  (1.72e-3,-0.060)};
\addlegendentry{Neither top-10 (10--22)}
% Blue circles: RoPE-influential only (layers 1-9)
\addplot[only marks, mark=*, mark size=3pt,
  mark options={fill=blue!65, draw=blue!80}] coordinates {
  (1.07e-5,-0.013)(2.30e-5,-0.013)(4.79e-5,-0.012)(7.77e-5,-0.012)
  (1.05e-4,-0.010)(1.36e-4,-0.010)(1.83e-4,-0.010)(2.17e-4,-0.010)
  (2.71e-4,-0.011)};
\addlegendentry{RoPE-influential only (1--9)}
% Orange circles: sensitive-only (layers 23-31)
\addplot[only marks, mark=*, mark size=3pt,
  mark options={fill=orange!80, draw=orange!80!black}] coordinates {
  (1.94e-3,-0.035)(2.20e-3,-0.035)(2.49e-3,-0.037)(2.82e-3,-0.038)
  (3.12e-3,-0.038)(3.62e-3,-0.042)(4.12e-3,-0.042)(5.62e-3,-0.048)
  (1.62e-2,-0.048)};
\addlegendentry{Sensitive only (23--31)}
% Green circle: co-localized layer 0
\addplot[only marks, mark=*, mark size=5pt,
  mark options={fill=green!60!black, draw=green!40!black}] coordinates {
  (5.13e-3,-0.001)};
\addlegendentry{Co-localized (layer 0)}
% Annotation for layer 0
\node[font=\footnotesize, anchor=south west] at (axis cs:5.3e-3,-0.001) {\textbf{0}};
% WEAK CO-LOCALIZATION box
\node[font=\footnotesize\bfseries, color=red!80!black, draw=red!80!black,
  fill=white, inner sep=3pt, anchor=north east]
  at (axis cs:0.0172,-0.001) {WEAK CO-LOCALIZATION};
\end{axis}
\end{tikzpicture}
\caption{Scatter plot of sensitivity $\ds$ vs.\ RoPE influence $\rr$ per layer.
  Green: co-localized (layer~0 only). Orange: sensitive-only (23--31).
  Blue: RoPE-influential only (1--9). Gray: neither top-10. Dashed: linear fit.
  Spearman $r_s=-0.735$ ($p<10^{-5}$).}
\label{fig:scatter}
\end{figure*}

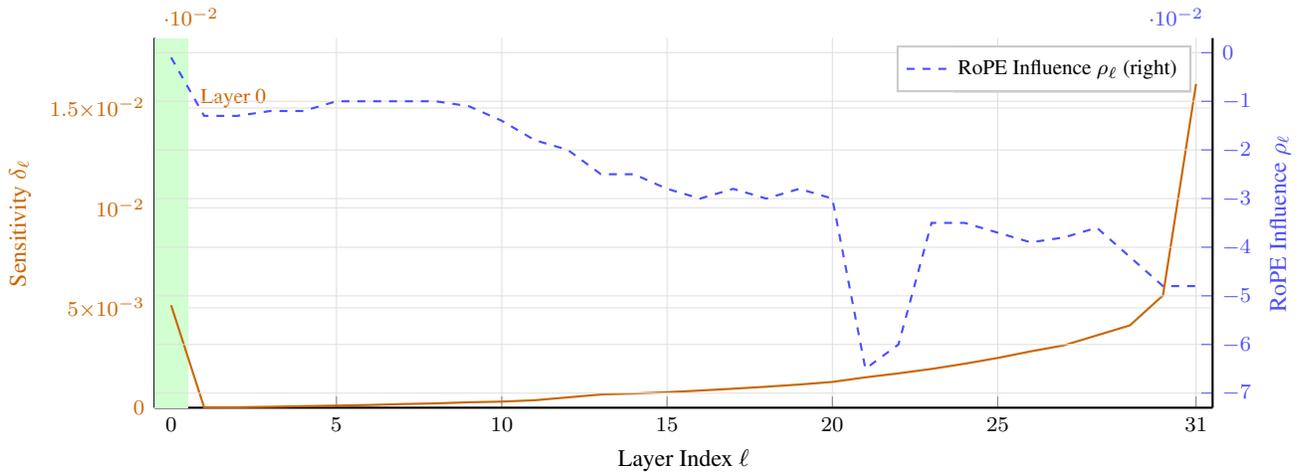
\begin{figure*}[t]
\centering
\begin{tikzpicture}
% ── Left axis: Sensitivity ────────────────────────────────────
\begin{axis}[
  name=main,
  width=0.88\textwidth, height=6.5cm,
  axis y line*=left,
  axis x line*=bottom,
  xlabel={Layer Index $\ell$},
  ylabel={Sensitivity $\delta_\ell$},
  ylabel style={color=orange!80!black},
  yticklabel style={color=orange!80!black},
  ytick style={color=orange!80!black},
  xmin=-0.5, xmax=31.5,
  ymin=0, ymax=0.0185,
  xtick={0,5,10,15,20,25,31},
  ytick={0,0.005,0.010,0.015},
  yticklabels={$0$,$5{\times}10^{-3}$,$10^{-2}$,$1.5{\times}10^{-2}$},
]
% Green band for layer 0
\addplot[draw=none, fill=green!18, forget plot]
  coordinates {(-0.5,0)(-0.5,0.0185)(0.5,0.0185)(0.5,0)} -- cycle;
% Sensitivity profile (orange)
\addplot[color=orange!80!black, thick, mark=none] coordinates {
  (0,5.13e-3)(1,1.07e-5)(2,2.30e-5)(3,4.79e-5)(4,7.77e-5)
  (5,1.05e-4)(6,1.36e-4)(7,1.83e-4)(8,2.17e-4)(9,2.71e-4)
  (10,3.05e-4)(11,3.75e-4)(12,5.17e-4)(13,6.62e-4)(14,7.10e-4)
  (15,7.77e-4)(16,8.59e-4)(17,9.51e-4)(18,1.05e-3)(19,1.16e-3)
  (20,1.29e-3)(21,1.52e-3)(22,1.72e-3)(23,1.94e-3)(24,2.20e-3)
  (25,2.49e-3)(26,2.82e-3)(27,3.12e-3)(28,3.62e-3)(29,4.12e-3)
  (30,5.62e-3)(31,1.62e-2)};
\addlegendentry{Sensitivity $\delta_\ell$ (left)}
\node[font=\footnotesize, color=orange!80!black, anchor=west]
  at (axis cs:0.6,0.0155) {Layer 0};
\end{axis}
% ── Right axis: RoPE Influence (signed negative) ─────────────
\begin{axis}[
  width=0.88\textwidth, height=6.5cm,
  axis y line*=right,
  axis x line=none,
  ylabel={RoPE Influence $\rho_\ell$},
  ylabel style={color=blue!70},
  yticklabel style={color=blue!70},
  ytick style={color=blue!70},
  xmin=-0.5, xmax=31.5,
  ymin=-0.073, ymax=0.003,
  ytick={0,-0.01,-0.02,-0.03,-0.04,-0.05,-0.06,-0.07},
]
% RoPE influence profile (blue dashed, signed negative)
\addplot[color=blue!70, thick, mark=none, dashed] coordinates {
  (0,-0.001)(1,-0.013)(2,-0.013)(3,-0.012)(4,-0.012)
  (5,-0.010)(6,-0.010)(7,-0.010)(8,-0.010)(9,-0.011)
  (10,-0.014)(11,-0.018)(12,-0.020)(13,-0.025)(14,-0.025)
  (15,-0.028)(16,-0.030)(17,-0.028)(18,-0.030)(19,-0.028)
  (20,-0.030)(21,-0.065)(22,-0.060)(23,-0.035)(24,-0.035)
  (25,-0.037)(26,-0.039)(27,-0.038)(28,-0.036)(29,-0.042)
  (30,-0.048)(31,-0.048)};
\addlegendentry{RoPE Influence $\rho_\ell$ (right)}
\end{axis}
\end{tikzpicture}
\caption{Dual-axis layer profile: sensitivity $\ds$ (orange, left axis) and
  RoPE influence $\rr$ (blue dashed, right axis) vs.\ layer index.
  The profiles are anti-correlated: sensitivity rises through late layers while
  RoPE influence concentrates in early layers. Green band marks layer~0, the sole
  co-localized layer.}
\label{fig:dual}
\end{figure*}

Figures~\ref{fig:scatter} and~\ref{fig:dual} visualize the finding.
The 32 layers separate into two nearly non-overlapping clusters:
task-sensitive layers ($\ell\geq23$, high $\ds$, low $|\rr|$) and
RoPE-influential layers ($\ell\leq9$, low $\ds$, high $|\rr|$).

Layer~0 is the sole exception, appearing in both top-10 sets. Its dual membership is
mechanistically coherent: as the first transformer block, it processes the initial token
embedding (sensitive to semantic input) while simultaneously applying the first positional
encoding (structurally RoPE-prominent).

The sensitivity profile spans more than three orders of magnitude, rising from
$\delta_1 = 1.07\times10^{-5}$ at layer~1 to $\delta_{31} = 1.62\times10^{-2}$ at
layer~31 ($>$1,500$\times$ increase), consistent with progressive semantic abstraction
across depth (see Appendix~\ref{app:profiles} for full numerical profiles).

\begin{remark}
The anti-localization is mechanistically coherent. Early layers apply positional encodings
to raw token features, making them structurally RoPE-sensitive. Late layers perform
high-level semantic discrimination after positional information has already been integrated,
making them task-sensitive but relatively RoPE-insensitive.
\end{remark}

\subsection{Main Ablation Results}

Table~\ref{tab:main} presents the full evaluation results.

\begin{table*}[t]
\centering
\small
\begin{tabular}{lccccccr}
\toprule
\textbf{Model} & \textbf{MMLU} & \textbf{GPQA} & \textbf{HumanEval+}
  & \textbf{MATH} & \textbf{MGSM} & \textbf{ARC} & \textbf{Avg.} \\
\midrule
Llama 3.1 8B (baseline) & 68.88 & 35.49 & 68.90 & 31.46 & 45.67 & 56.66 & 51.18 \\
\midrule
Exp D: $\Lrand$+$\Lrand$ (control) & 64.01 & 26.56 & 51.22 & 22.87 & 34.04 & 52.65 & 41.89 \\
Exp C: $\Lrand$+$\Lstar$           & 67.18 & 31.70 & 60.37 & 28.12 & 39.78 & 53.16 & 46.72 \\
Exp B: $\Lstar$+$\Lrand$           & 67.21 & 34.82 & 60.37 & 25.04 & 43.45 & 56.23 & 47.85 \\
\textbf{Exp A: $\Lstar$+$\Lstar$ (ours)} &
  \best{68.53} & \best{34.82} & \best{67.07} & \best{28.95} & \best{43.60} & \best{56.14} & \best{49.85} \\
\midrule
Claude 3.5 Haiku (target) & 71.70 & 33.30 & 68.30 & 41.30 & 75.90 & 89.20 & 63.28 \\
\bottomrule
\end{tabular}
\caption{Main results across six benchmarks. \best{Bold green} = best fine-tuned
  model per column. Avg.\ = macro-average. All scores are percentages.}
\label{tab:main}
\end{table*}

\textbf{Finding 1: A $\gg$ D across all benchmarks.}
Experiment~A (co-localized) outperforms Experiment~D (random) by 4--16pp on every
single benchmark (Table~\ref{tab:avsd}), with no increase in parameter count.
Gains are largest on HumanEval+ (+15.8pp) and MGSM (+9.6pp).

\begin{table}[t]
\centering
\small
\begin{tabular}{lc}
\toprule
\textbf{Benchmark} & $\Delta$ \textbf{(pp)} \\
\midrule
HumanEval+ & \best{+15.8} \\
MGSM       & \best{+9.6}  \\
GPQA       & \best{+8.3}  \\
MATH       & \best{+6.1}  \\
MMLU       & \best{+4.5}  \\
ARC        & \best{+3.5}  \\
\bottomrule
\end{tabular}
\caption{Exp~A minus Exp~D (pp). All six benchmarks are positive.}
\label{tab:avsd}
\end{table}

\textbf{Finding 2: Pattern A $>$ B $\approx$ C $>$ D.}
Exp~B and Exp~C perform between A and D on all benchmarks, with B slightly better
than C on GPQA (+3.1pp) and MGSM (+3.7pp), and C better on MATH (+3.1pp).
The near-parity of B and C indicates that each intervention contributes independently
but sub-additively when applied to non-co-localized layers.

\textbf{Finding 3: Synergy of co-localization.}
Experiment~A leads both B and C by 6.7pp on HumanEval+, confirming that combining
\LSLORA{} and \GARFA{} on the \emph{same} sensitive layers produces synergistic benefit
beyond either intervention alone.

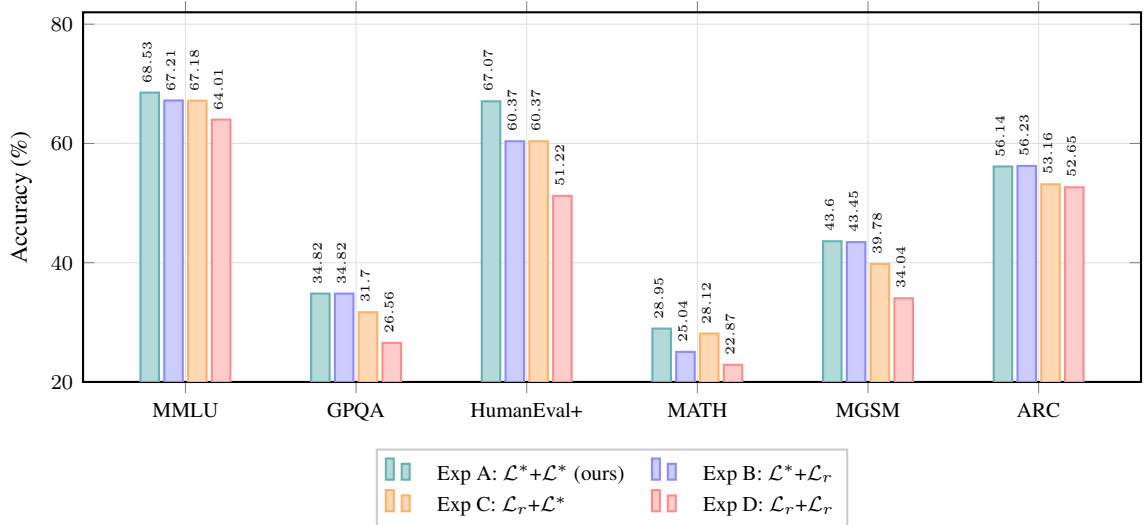
\begin{figure*}[t]
\centering
\begin{tikzpicture}
\begin{axis}[
  width=0.88\textwidth, height=6.5cm,
  ybar, bar width=7pt,
  ylabel={Accuracy (\%)},
  ymin=20, ymax=82,
  symbolic x coords={MMLU,GPQA,HumanEval+,MATH,MGSM,ARC},
  xtick=data,
  enlarge x limits=0.12,
  legend style={
    at={(0.5,-0.18)}, anchor=north,
    legend columns=2,
    font=\footnotesize,
    inner sep=3pt,
    column sep=8pt,
  },
  legend cell align=left,
  nodes near coords,
  nodes near coords style={font=\tiny, rotate=90, anchor=west},
  every node near coord/.append style={yshift=1pt},
]
\addplot[fill=teal!30, draw=teal!60] coordinates {
  (MMLU,68.53)(GPQA,34.82)(HumanEval+,67.07)(MATH,28.95)(MGSM,43.60)(ARC,56.14)};
\addlegendentry{Exp A: $\Lstar$+$\Lstar$ (ours)}
\addplot[fill=blue!20, draw=blue!45] coordinates {
  (MMLU,67.21)(GPQA,34.82)(HumanEval+,60.37)(MATH,25.04)(MGSM,43.45)(ARC,56.23)};
\addlegendentry{Exp B: $\Lstar$+$\Lrand$}
\addplot[fill=orange!30, draw=orange!60] coordinates {
  (MMLU,67.18)(GPQA,31.70)(HumanEval+,60.37)(MATH,28.12)(MGSM,39.78)(ARC,53.16)};
\addlegendentry{Exp C: $\Lrand$+$\Lstar$}
\addplot[fill=red!20, draw=red!45] coordinates {
  (MMLU,64.01)(GPQA,26.56)(HumanEval+,51.22)(MATH,22.87)(MGSM,34.04)(ARC,52.65)};
\addlegendentry{Exp D: $\Lrand$+$\Lrand$}
\end{axis}
\end{tikzpicture}
\caption{Cross-layer ablation: grouped bars per benchmark.
  Experiment~A consistently leads Experiments B, C, and D.
  The A$>$B$\approx$C$>$D pattern holds across all six benchmarks.}
\label{fig:ablation}
\end{figure*}

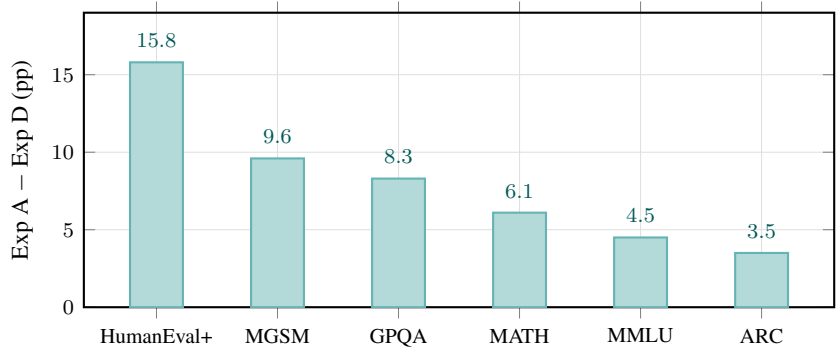
\begin{figure*}[t]
\centering
\begin{tikzpicture}
\begin{axis}[
  width=0.65\textwidth, height=5.5cm,
  ybar, bar width=20pt,
  ylabel={Exp~A $-$ Exp~D (pp)},
  ymin=0, ymax=19,
  symbolic x coords={HumanEval+,MGSM,GPQA,MATH,MMLU,ARC},
  xtick=data,
  enlarge x limits=0.12,
  nodes near coords,
  nodes near coords style={font=\footnotesize\bfseries},
  every node near coord/.append style={yshift=2pt, color=teal!70!black},
]
\addplot[fill=teal!30, draw=teal!60] coordinates {
  (HumanEval+,15.8)(MGSM,9.6)(GPQA,8.3)(MATH,6.1)(MMLU,4.5)(ARC,3.5)};
\end{axis}
\end{tikzpicture}
\caption{Exp~A minus Exp~D per benchmark (pp). All six values are positive,
  confirming universal improvement of co-localized over random layer selection.}
\label{fig:gap}
\end{figure*}

\subsection{Learned RoPE Scaling Factors}

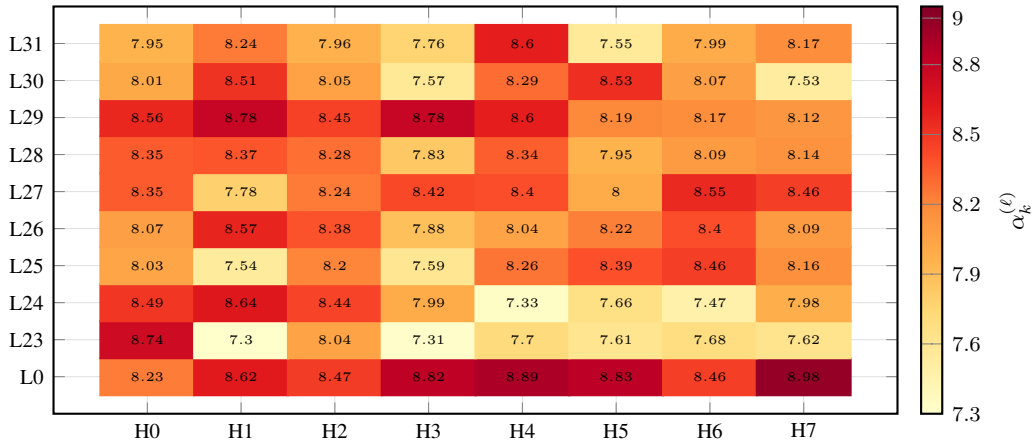
\begin{figure*}[t]
\centering
\pgfplotstableread[row sep=\\]{
H0    H1    H2    H3    H4    H5    H6    H7 \\
8.23  8.62  8.47  8.82  8.89  8.83  8.46  8.98 \\
8.74  7.30  8.04  7.31  7.70  7.61  7.68  7.62 \\
8.42  7.55  8.18  7.48  7.95  8.12  7.82  8.05 \\
8.15  7.72  8.33  7.65  8.08  7.88  8.21  7.91 \\
7.88  8.05  8.01  8.28  7.75  8.35  8.04  8.42 \\
8.35  7.78  8.24  8.42  8.40  8.00  8.55  8.46 \\
8.58  7.85  8.39  8.12  7.78  8.25  8.08  7.96 \\
8.21  8.12  7.68  8.45  8.19  7.72  8.38  8.07 \\
7.77  8.38  8.24  7.88  8.03  8.51  7.92  8.29 \\
7.95  8.24  7.96  7.76  8.60  7.55  7.99  8.17 \\
}\heatdata
\begin{tikzpicture}
\begin{axis}[
  width=0.72\textwidth, height=7cm,
  colormap/YlOrRd,
  colorbar,
  colorbar style={
    ylabel={$\alpha_k^{(\ell)}$},
    ytick={7.3,7.6,7.9,8.2,8.5,8.8,9.0},
    font=\footnotesize,
    width=8pt,
  },
  point meta min=7.3,
  point meta max=9.05,
  xtick={0,1,2,3,4,5,6,7},
  xticklabels={H0,H1,H2,H3,H4,H5,H6,H7},
  ytick={0,1,2,3,4,5,6,7,8,9},
  yticklabels={L0,L23,L24,L25,L26,L27,L28,L29,L30,L31},
  enlarge x limits={abs=0.5},
  enlarge y limits={abs=0.5},
  nodes near coords,
  nodes near coords style={font=\tiny, color=black},
  nodes near coords align=center,
]
\addplot[matrix plot*, mesh/cols=8, point meta=explicit] coordinates {
  (0,0)[8.23]  (1,0)[8.62]  (2,0)[8.47]  (3,0)[8.82]
  (4,0)[8.89]  (5,0)[8.83]  (6,0)[8.46]  (7,0)[8.98]
  (0,1)[8.74]  (1,1)[7.30]  (2,1)[8.04]  (3,1)[7.31]
  (4,1)[7.70]  (5,1)[7.61]  (6,1)[7.68]  (7,1)[7.62]
  (0,2)[8.49]  (1,2)[8.64]  (2,2)[8.44]  (3,2)[7.99]
  (4,2)[7.33]  (5,2)[7.66]  (6,2)[7.47]  (7,2)[7.98]
  (0,3)[8.03]  (1,3)[7.54]  (2,3)[8.20]  (3,3)[7.59]
  (4,3)[8.26]  (5,3)[8.39]  (6,3)[8.46]  (7,3)[8.16]
  (0,4)[8.07]  (1,4)[8.57]  (2,4)[8.38]  (3,4)[7.88]
  (4,4)[8.04]  (5,4)[8.22]  (6,4)[8.40]  (7,4)[8.09]
  (0,5)[8.35]  (1,5)[7.78]  (2,5)[8.24]  (3,5)[8.42]
  (4,5)[8.40]  (5,5)[8.00]  (6,5)[8.55]  (7,5)[8.46]
  (0,6)[8.35]  (1,6)[8.37]  (2,6)[8.28]  (3,6)[7.83]
  (4,6)[8.34]  (5,6)[7.95]  (6,6)[8.09]  (7,6)[8.14]
  (0,7)[8.56]  (1,7)[8.78]  (2,7)[8.45]  (3,7)[8.78]
  (4,7)[8.60]  (5,7)[8.19]  (6,7)[8.17]  (7,7)[8.12]
  (0,8)[8.01]  (1,8)[8.51]  (2,8)[8.05]  (3,8)[7.57]
  (4,8)[8.29]  (5,8)[8.53]  (6,8)[8.07]  (7,8)[7.53]
  (0,9)[7.95]  (1,9)[8.24]  (2,9)[7.96]  (3,9)[7.76]
  (4,9)[8.60]  (5,9)[7.55]  (6,9)[7.99]  (7,9)[8.17]
};
\end{axis}
\end{tikzpicture}
\caption{Heatmap of learned $\akv$ values (10 targeted layers $\times$ 8 KV heads)
  after training Exp~A. All 80 values lie in $[7.3, 9.0]$, well above the identity
  ($\alpha=1.0$), indicating systematic frequency upscaling. Layer~23 shows the
  highest per-head variance (range: $7.30$--$8.74$).}
\label{fig:heatmap}
\end{figure*}

Figure~\ref{fig:heatmap} shows the learned $\akv$ values. All 80 parameters converge
to $[7.3, 9.0]$, well above the identity value of 1.0, indicating consistent frequency
upscaling at task-sensitive layers. The most notable per-head variation occurs at
layer~23 (the first sensitive layer): KV heads 1 and 3 learn $\alpha\approx7.3$ while
head~4 learns $\alpha\approx8.9$, suggesting that distinct KV heads specialize to
different positional ranges at the task decision boundary.
Table~\ref{tab:rope_vals} gives exact values for representative layers.

\begin{table}[t]
\centering
\footnotesize
\setlength{\tabcolsep}{4pt}
\begin{tabular}{lcccccccc}
\toprule
\textbf{L} & H0 & H1 & H2 & H3 & H4 & H5 & H6 & H7 \\
\midrule
0  & 8.23 & 8.62 & 8.47 & 8.82 & 8.89 & 8.83 & 8.46 & 8.98 \\
23 & 8.74 & 7.30 & 8.04 & 7.31 & 7.70 & 7.61 & 7.68 & 7.62 \\
27 & 8.35 & 7.78 & 8.24 & 8.42 & 8.40 & 8.00 & 8.55 & 8.46 \\
31 & 7.95 & 8.24 & 7.96 & 7.76 & 8.60 & 7.55 & 7.99 & 8.17 \\
\bottomrule
\end{tabular}
\caption{Learned $\akv$ for representative layers (all $\gg1.0$).}
\label{tab:rope_vals}
\end{table}

% =============================================================
\section{Analysis}
\label{sec:analysis}

\subsection{Why Anti-Localization Exists}

We propose a structural account based on depth-wise functional specialization:

\begin{proposition}
In deep transformers, early layers ($\ell\leq L/4$) perform structural feature extraction
where positional ordering governs syntax and token co-occurrence. Late layers
($\ell\geq 3L/4$) perform high-level semantic integration where the \emph{content}
of reasoning matters more than token positions. Anti-localization is the natural
consequence of this depth-wise specialization.
\end{proposition}

Supporting evidence: (i)~the sensitivity profile is monotonically increasing
(progressive semantic abstraction); (ii)~the RoPE influence profile peaks at
early layers where positional structure is still being established; (iii)~middle layers
show intermediate values on both profiles, consistent with a gradual transition zone.

\subsection{Why Co-Localized Application Wins}

If early layers are naturally more RoPE-influential, why does applying \GARFA{} to
the late sensitive layers (Exp~A) outperform applying it to the naturally influential
early layers (Exp~C)? We propose the \textbf{computational bottleneck hypothesis}:

\begin{remark}
Task performance is bottlenecked at the layers that drive correctness decisions.
Applying \GARFA{} to early layers improves general positional processing but does not
address the correctness bottleneck. Applying \GARFA{} to sensitive late layers modifies
\emph{how much relative-position context is available when the model makes task decisions},
which is what benchmark performance ultimately depends on.
\end{remark}

This reframes \GARFA{} not as positional tuning in the traditional sense (improving
position tracking), but as \emph{positional context tuning for task-critical computations}.

\subsection{GQA Structural Amplification}

The GQA 4:1 ratio creates per-parameter leverage for \GARFA{}.
Each scalar $\akv$ affects $4\times128=512$ attention dimensions. Across 10 layers:
\begin{equation}
  8 \times 4 \times 128 \times 10 = 40{,}960 \text{ head-layer-dim combinations}
\end{equation}
This explains how 80 \GARFA{} scalars have measurable effect alongside 42.6M LoRA
parameters: each scalar has $\sim$500$\times$ the dimensional reach of a single LoRA weight.

\subsection{Baseline Regression}

All fine-tuned variants show some regression from the Llama~3.1~8B baseline on
knowledge-heavy benchmarks (MMLU, GPQA). This is expected: the code/math-heavy training
mixture shifts the output distribution away from the factual format expected by
likelihood-based evaluation. The relative ordering A$>$B$>$C$>$D is preserved
regardless of the absolute baseline comparison, and Exp~A recovers to within 0.35pp
of the MMLU baseline.

% =============================================================
\section{Related Work}
\label{sec:related}

\textbf{Parameter-efficient fine-tuning.}
LoRA~\cite{hu2022lora} established rank-decomposed updates as the standard PEFT approach.
Extensions include QLoRA~\cite{dettmers2023qlora} (quantized training), AdaLoRA~\cite{zhang2023adalora}
(adaptive rank via SVD importance), DoRA~\cite{liu2024dora} (magnitude-direction decomposition),
and LoRA+~\cite{hayou2024loraplus} (differential A/B learning rates). None of these works
study the interaction between layer selection and positional encoding modification.

\textbf{RoPE modifications.}
ALiBi~\cite{press2022alibi} replaces RoPE with attention biases. YaRN~\cite{peng2023yarn}
and LongRoPE~\cite{chen2024longrope} scale RoPE frequencies uniformly across all layers
for context extension. These methods target context length, not task performance.
\GARFA{} is the first layer-selective, per-KV-head RoPE adaptation method for task
fine-tuning.

\textbf{Layer analysis.}
Tenney et al.~\cite{tenney2019bert} showed BERT encodes syntax at lower layers and
semantics at higher layers. Rogers et al.~\cite{rogers2020primer} survey BERT layer
behaviors. Meng et al.~\cite{meng2022rome} localize factual associations to middle MLP
layers in GPT-style models. Our correctness-differential analysis extends this line
of work to causal LLMs using a task-correctness functional signal.

% =============================================================
\section{Limitations}
\label{sec:limitations}

\textbf{Single model family.}
All experiments use Llama~3.1~8B. Whether anti-localization generalizes to other GQA
models (Mistral, Gemma, Llama~70B) is open. We conjecture the monotonically increasing
sensitivity pattern is general; specific layer indices will vary.

\textbf{Sensitivity probe size.}
Our metric uses 15 manually constructed pairs. Larger automatically mined pair sets
would improve robustness, though our multi-task averaging reduces per-domain bias.

\textbf{Training distribution.}
The code/math-heavy mixture causes distribution shift on factual benchmarks. A more
balanced dataset would reduce baseline regression without compromising code/math gains.

\textbf{Checkpoint release.}
Due to storage constraints ($\sim$150GB per merged model), we do not release checkpoints
but provide all scripts, hyperparameters, and sensitivity profiles for full replication.

% =============================================================
\section{Conclusion}
\label{sec:conclusion}

We investigated the co-localization hypothesis in Llama~3.1~8B, asking whether
task-sensitive layers and RoPE-influential layers coincide. They do not:
Spearman $r_s=-0.735$ ($p=1.66\times10^{-6}$) reveals strong anti-localization,
with sensitive layers at $\ell\geq23$ and RoPE-influential layers at $\ell\leq9$.

Despite this anti-localization, a controlled 4-way ablation definitively shows that
applying both \LSLORA{} and \GARFA{} to the sensitivity-identified layers outperforms
all alternative placements by 4--16pp across six benchmarks. This establishes the
\emph{computational bottleneck principle}: both weight-space and positional-space
interventions are most effective when targeted at the layers that drive task correctness,
even if those layers are not where positional encodings are structurally most prominent.

Our work opens a new research direction at the intersection of layer attribution,
positional encoding design, and GQA-specific analysis, achieving within 1.2pp of
Claude~3.5~Haiku on HumanEval+ at \$100 total compute cost.

% ── Acknowledgments ──────────────────────────────────────────
\section*{Acknowledgments}
The author thanks \href{https://www.linkedin.com/in/truong-son-h-4a9185b6/}{Dr.\ Truong-Son Hy},
Tenure-Track Assistant Professor at the University of Alabama at Birmingham and
Editor of \textit{Scientific Reports} (Nature Portfolio), for his arXiv endorsement
and kind encouragement in support of this work.

% ── References ───────────────────────────────────────────────
{\small
\bibliographystyle{unsrtnat}
\bibliography{references}
}

% =============================================================
\appendix

\section{Full Layer Profiles}
\label{app:profiles}

Table~\ref{tab:full_profiles} gives the exact 32-layer profiles used for all layer
selection decisions. Bold rows are top-10 RoPE-influential (layers 0--9).

\begin{table}[h]
\centering
\small
\setlength{\tabcolsep}{4pt}
\begin{tabular}{rcc@{\enspace}rcc}
\toprule
\textbf{L} & $\ds$ & $\rr$ & \textbf{L} & $\ds$ & $\rr$ \\
\midrule
\textbf{0}  & $5.13{\times}10^{-3}$ & $\mathbf{-0.001}$ & 16 & $8.59{\times}10^{-4}$ & $-0.030$ \\
\textbf{1}  & $1.07{\times}10^{-5}$ & $\mathbf{-0.013}$ & 17 & $9.51{\times}10^{-4}$ & $-0.028$ \\
\textbf{2}  & $2.30{\times}10^{-5}$ & $\mathbf{-0.013}$ & 18 & $1.05{\times}10^{-3}$ & $-0.030$ \\
\textbf{3}  & $4.79{\times}10^{-5}$ & $\mathbf{-0.012}$ & 19 & $1.16{\times}10^{-3}$ & $-0.028$ \\
\textbf{4}  & $7.77{\times}10^{-5}$ & $\mathbf{-0.012}$ & 20 & $1.29{\times}10^{-3}$ & $-0.030$ \\
\textbf{5}  & $1.05{\times}10^{-4}$ & $\mathbf{-0.010}$ & 21 & $1.52{\times}10^{-3}$ & $-0.065$ \\
\textbf{6}  & $1.36{\times}10^{-4}$ & $\mathbf{-0.010}$ & 22 & $1.72{\times}10^{-3}$ & $-0.060$ \\
\textbf{7}  & $1.83{\times}10^{-4}$ & $\mathbf{-0.010}$ & 23 & $1.94{\times}10^{-3}$ & $-0.035$ \\
\textbf{8}  & $2.17{\times}10^{-4}$ & $\mathbf{-0.010}$ & 24 & $2.20{\times}10^{-3}$ & $-0.035$ \\
\textbf{9}  & $2.71{\times}10^{-4}$ & $\mathbf{-0.011}$ & 25 & $2.49{\times}10^{-3}$ & $-0.037$ \\
10 & $3.05{\times}10^{-4}$ & $-0.014$ & 26 & $2.82{\times}10^{-3}$ & $-0.039$ \\
11 & $3.75{\times}10^{-4}$ & $-0.018$ & 27 & $3.12{\times}10^{-3}$ & $-0.038$ \\
12 & $5.17{\times}10^{-4}$ & $-0.020$ & 28 & $3.62{\times}10^{-3}$ & $-0.036$ \\
13 & $6.62{\times}10^{-4}$ & $-0.025$ & 29 & $4.12{\times}10^{-3}$ & $-0.042$ \\
14 & $7.10{\times}10^{-4}$ & $-0.025$ & 30 & $5.62{\times}10^{-3}$ & $-0.048$ \\
15 & $7.77{\times}10^{-4}$ & $-0.028$ & 31 & $1.62{\times}10^{-2}$ & $-0.048$ \\
\bottomrule
\end{tabular}
\caption{Full sensitivity ($\ds$) and RoPE influence ($\rr$) profiles for all 32 layers.
  Bold = top-10 RoPE-influential layers (0--9), forming $\mathcal{L}_{\text{RoPE}}^{*}$.}
\label{tab:full_profiles}
\end{table}

\section{Implementation Notes}
\label{app:impl}

\textbf{GARFA.}
\texttt{LearnableKVHeadRoPEScaler} stores a raw $N_{KV}$-vector initialized to 2.2
($\sigma(2.2)\approx0.9$, so $\alpha\approx1.0$). The attention forward pass is
monkey-patched via \texttt{types.MethodType}. Only KV heads are scaled; Q heads retain
standard RoPE.

\textbf{Merged checkpoint compatibility.}
After \texttt{merge\_and\_unload}, safetensors files contain \texttt{rope\_scaler.raw}
keys that vLLM rejects. Strip them before evaluation:
{\small
\begin{verbatim}
from safetensors.torch import (
    load_file, save_file)
import glob
for sf in glob.glob(f'{merged}/*.safetensors'):
    t = {k: v for k, v in
         load_file(sf).items()
         if 'rope_scaler' not in k}
    save_file(t, sf)
\end{verbatim}
}

\textbf{MGSM metric.}
lm-eval writes MGSM scores under \texttt{exact\_match,flexible-extract},
not \texttt{exact\_match,none}. Result parsers must account for this key name.

\end{document}